\documentclass[letterpaper,twocolumn,10pt]{article}
\usepackage{usenix,epsfig,endnotes}
\usepackage{graphicx}
\usepackage{subcaption}
\usepackage{setspace}
\usepackage{float}
\usepackage{amsmath}
\microtypecontext{spacing=nonfrench}

\begin{document}

\date{}

\title{\Large \bf Evaluating Zero-Shot Long-Context LLM Compression 
}

\author{
{\rm Chenyu Wang}\\
cw9420@princeton.edu
\and
{\rm Yihan Wang}\\
yihanw\_mems@princeton.edu
\and
{\rm Kai Li}\\
li@cs.princeton.edu
}

\maketitle

\thispagestyle{empty}

\subsection*{Abstract}
This study evaluates the effectiveness of zero-shot compression techniques on large language models (LLMs) under long-context. We identify the tendency for computational errors to increase under long-context when employing certain compression methods. We propose a hypothesis to explain the varied behavior of different LLM compression techniques and explore remedies to mitigate the performance decline observed in some techniques under long-context. This is a course report for COS 598D Machine Learning and Systems by Prof. Kai Li at Princeton University. Due to limited computational resources, our experiments were conducted only on LLaMA-2-7B-32K. 
\section{Introduction}

Large Language Models (LLMs) excel in processing and generating human-like text, yet they confront significant challenges.\cite{jiang2023mistral, gpt3, touvron2023llama2, openai2023gpt4, black2022gpt, zhang2023llama} One major challenge is computational efficiency; LLMs demand considerable resources during both training and inference stages. Another challenge is their limited context length, which is the maximum number of tokens the model can process at once. Inputs that exceed this context length can result in unreasonable responses from the LLM.

To address the computational efficiency challenge of LLMs, many researchers have proposed leveraging model compression techniques. These LLM compression techniques commonly operate under the assumption that both the weights and activations of LLMs can be compressed with minimal impact on computational error\cite{xiao2022smoothquant, sun2023simple, kang2024gear}. On the other hand, researchers are also working on extending the context length of LLMs\cite{longbench, jiang_longllmlingua_2023}. The context length of LLMs have been expanding exponentially in recent years. This significant increase in context length allows LLMs to handle increasingly complex tasks, such as analyzing multiple books and documents simultaneously.


After reviewing the literature, we have observed that the majority of studies on LLM compression focus on models with relatively short context lengths (e.g., 4K). However, the effectiveness of these compression techniques on models with extensive context lengths (e.g., 32K) remains under-evaluated. In this project, we aim to evaluate zero-shot LLM compression under long-context. We begin by conducting both qualitative theoretical analysis and empirical evaluations of long-context LLM compression. Subsequently, we attempt to develop appropriate solutions to address the performance decline observed in some LLM compression techniques under long-context.

The contribution of this project can be summarized as follows:

\begin{itemize}
\item We conduct a theoretical analysis of long-context LLM compression. We find that as the context length increases in compressed LLMs, computational errors tend to accumulate.
\item We perform an empirical evaluation of various LLM compression techniques to assess computational errors in extended contexts. Our findings indicate diverse behaviors across different compression methods.
\item We propose a hypothesis to explain the varied responses of different LLM compression techniques and explore remedies to mitigate the performance decline observed in some techniques under long-context scenarios.
\end{itemize}

The rest of this report is organized as follows. In Section~\ref{sec:related}, we will introduces previous research related to this project. In Section~\ref{sec:method}, we will expand the technical details of our project. In Section~\ref{sec:conclusion}, we will 
 conclude our project. In Section 5, we will introduce the future work. 

\section{Related Works}
\label{sec:related}
\paragraph{Long-Context LLMs} In recent times, there has been a push towards expanding the context length of Language Models (LLMs) efficiently through continuous pretraining or fine-tuning. One approach involves enhancing Rotary Position Embeddings (RoPE)~\cite{su2021roformer}, which has led to longer contexts of up to 128k ~\cite{chen2023extending,chen2024longlora,peng2024yarn}. Another line of research, exemplified by Mistral~\cite{jiang2023mistral}, introduces sliding window attention mechanisms that focus only on a portion of tokens from the preceding layer, thereby reducing computational demands and facilitating pretraining with longer contexts of up to 30k. However, due to the memory-intensive nature of autoregressive generation in LLMs~\cite{kwon2023efficient}, the storage of Key-Value (KV) caches for longer contexts slows down inference processes and necessitates GPUs with large VRAM capacities.

\paragraph{Quantization in LLMs}

Quantization is a commonly employed method for model compression. \cite{han2015learning, sun2022gibbon, wang2023epim} Researchers investigate two distinct settings for Large Language Model (LLM) quantization:

\begin{itemize}
    \item W8A8 quantization, where both activations and weights are quantized to INT8~\cite{dettmers2022llmint8, xiao2022smoothquant, wei2022outlier,wei2023outlier, sheng2023high}.
    \item Low-bit weight-only quantization (e.g., W4A16), where only weights are quantized into low-bit integers~\cite{frantar2022gptq,dettmers2022case,sheng_s-lora_2023,sheng2023high,park2022nuqmm}.
\end{itemize}

This work concentrates on the second setting, as it not only reduces the hardware requirements by necessitating a smaller memory size but also accelerates token generation, thus alleviating memory-bound workloads. Besides, we require the algorithm to be zero-shot, which is relatively low-cost and non-task-specific.

\paragraph{Pruning in LLMs}
Pruning, a well-established technique for compressing neural networks, involves eliminating weights to create sparse networks~\cite{lecun1989optimal}. It can be broadly classified into structured and unstructured approaches.

Structural pruning involves removing entire filters from the neural network, making it more conducive to hardware implementation. Various methods exist for implementing structural pruning, such as l1-dependent pruning~\cite{han2015learning,zafrir2021prune}, first-order importance estimation~\cite{lu2022learn}, hessian-based estimation~\cite{kurtic2022optimal,wang2019eigendamage}, or the optimal brain surgeon~\cite{lecun1989optimal,kurtic2022optimal}.

Unstructured methods~\cite{han2015learning, han2016deep,paul2022unmasking,sun2023simple} like magnitude pruning operate at the individual weight level, maintaining performance even at higher sparsity levels. However, existing pruning methods usually require modifications to the training procedure~\cite{sanh2021multitask}, retraining the pruned networks to regain accuracy~\cite{liu2019roberta}, or a computationally intensive iterative retraining process~\cite{bondarenko2021understanding,frankle2018lottery}. Yet, scaling these techniques to LLMs with billions of parameters poses a challenge, as the necessary training process demands significant computational resources~\cite{hoffmann2022training, zhang2022opt}.

We focus on unstructured pruning methods in this work as they are more fundamental and flexible than structural pruning. Specifically, we choose two representative methods: magnitude pruning and Wanda~\cite{sun2023simple}. Although other model compression methods such as neural architecture search exist~\cite{zoph2016neural, cai2022deepguiser}, this study focuses exclusively on pruning and quantization. 

\section{Concurrent Work}

In our review of the literature, we noted a related study on the quantization of long-context LLMs detailed in Li et al. (2024) \cite{li2024evaluating}. This research conducted two experiments to assess the performance of LLMs under extended contexts. It reported a decline in performance when applying weight-only quantization, weight-activation quantization, and KV cache quantization. However, it did not delve into further analysis to explore the underlying causes of this observed performance degradation.

\section{Evaluation Details}
\label{sec:method}
\subsection{Theoretical Analysis}

LLMs are typically built on transformer architecture, which operates as a 'sequence model'. In the transformer architecture, each newly generated token calculates its attention scores against the hidden states of all preceding tokens. Regarding LLM compression techniques, while they can accelerate inference, they also introduce computational errors in both the output and hidden states of LLMs. Consequently, in compressed LLMs, each new token is computed based on an increasingly large number of preceding tokens, with each token contributing its own computational error. The formulas below provide a detailed description of this process. 


Consider the transformer architecture processing a sequence where each token $t$ has an associated query vector $\mathbf{q}_t$ and key vectors $\mathbf{k}_1, \dots, \mathbf{k}_t$ for all previous tokens. The attention scores $\mathbf{a}_t$ for token $t$ are calculated as:
\begin{align}
    \textbf{a}_t = \text{softmax}\left(\frac{\textbf{q}_t [k_1^T \cdots k_{t}^T]}{\sqrt{d_k}}\right)
\end{align}
where $d_k$ is the dimensionality of the key vectors, ensuring proper scaling.

The hidden state $\mathbf{h}_t$ for token $t$ is computed by:
\begin{align}
    \mathbf{h}_t = [v_1^T, \cdots, v_t^T] \textbf{a}_t
\end{align}
where $\mathbf{v}_t$ represents the value vectors corresponding to token $t$.

When LLMs undergo compression, noise $\epsilon \sim \mathcal{N}(0, \sigma^2)$ is added to each key and value vector:

Then we have:
\begin{align}
    \mathbf{h}_t &= \sum_{i=1}^t \text{softmax}\left(\frac{\textbf{q}_i (k_i + \epsilon_{ki})}{\sqrt{d_k}}\right) (v_i + \epsilon_{vi})  \\ 
     &\approx \sum_{i=1}^t \text{softmax}\left(\frac{\textbf{q}_i k_i}{\sqrt{d_k}}\right) v_i + \\ & \sum_{i=1}^t (\text{softmax}\left(\frac{\textbf{q}_i k_i}{\sqrt{d_k}}\right) \epsilon_{vi} + \frac{\partial a_{ti}}{\partial k_i} v_i \epsilon_{ki})
\end{align}
for simplicity, in this step we assume that all the vectors involved are one dimensional.

\begin{figure*}[t]
    \centering
    \begin{subfigure}[b]{0.45\textwidth}
         \centering
         \includegraphics[width=\textwidth]{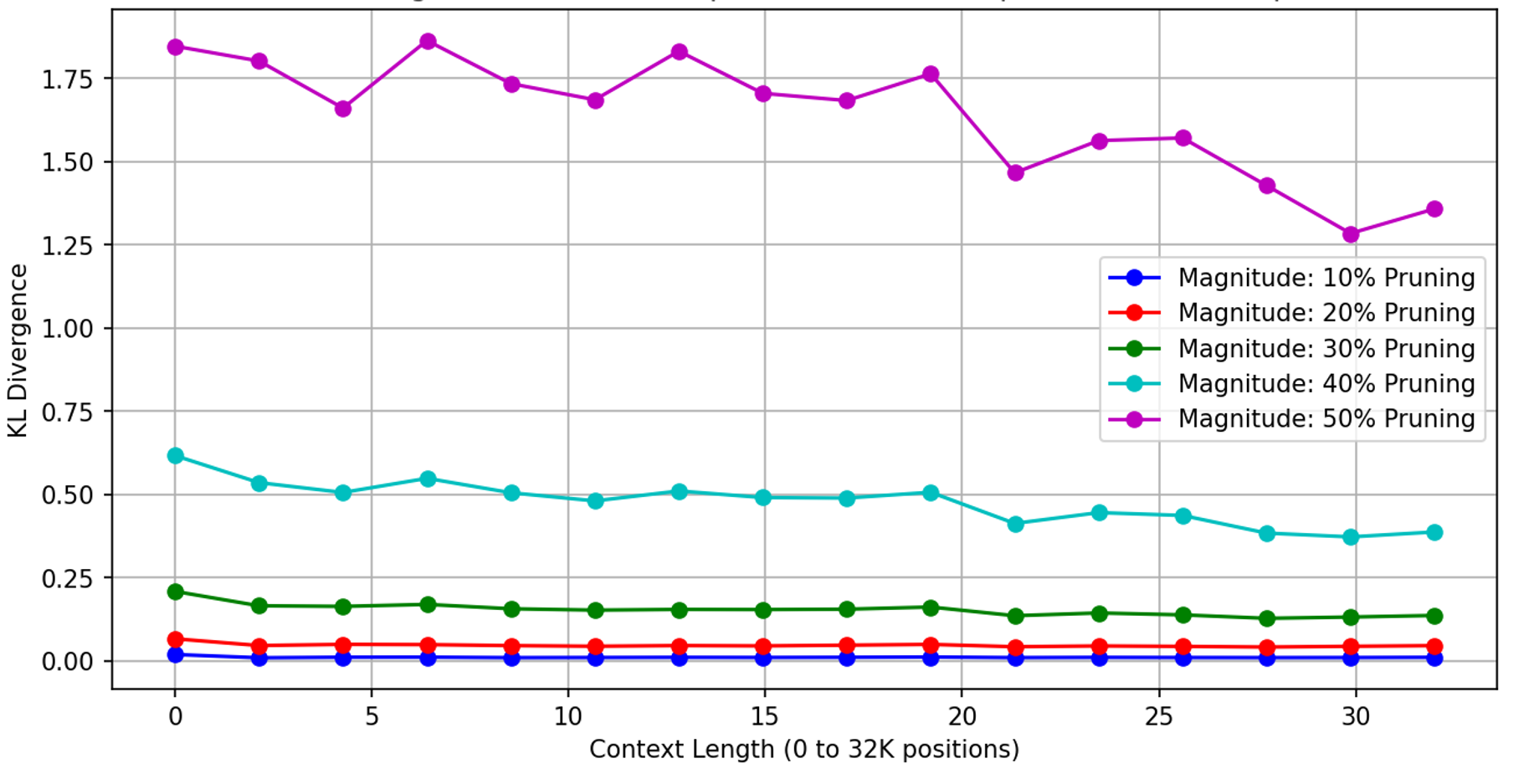}
    \end{subfigure}%
    ~ 
    \begin{subfigure}[t]{0.45\textwidth}
         \centering
         \includegraphics[width=\textwidth]{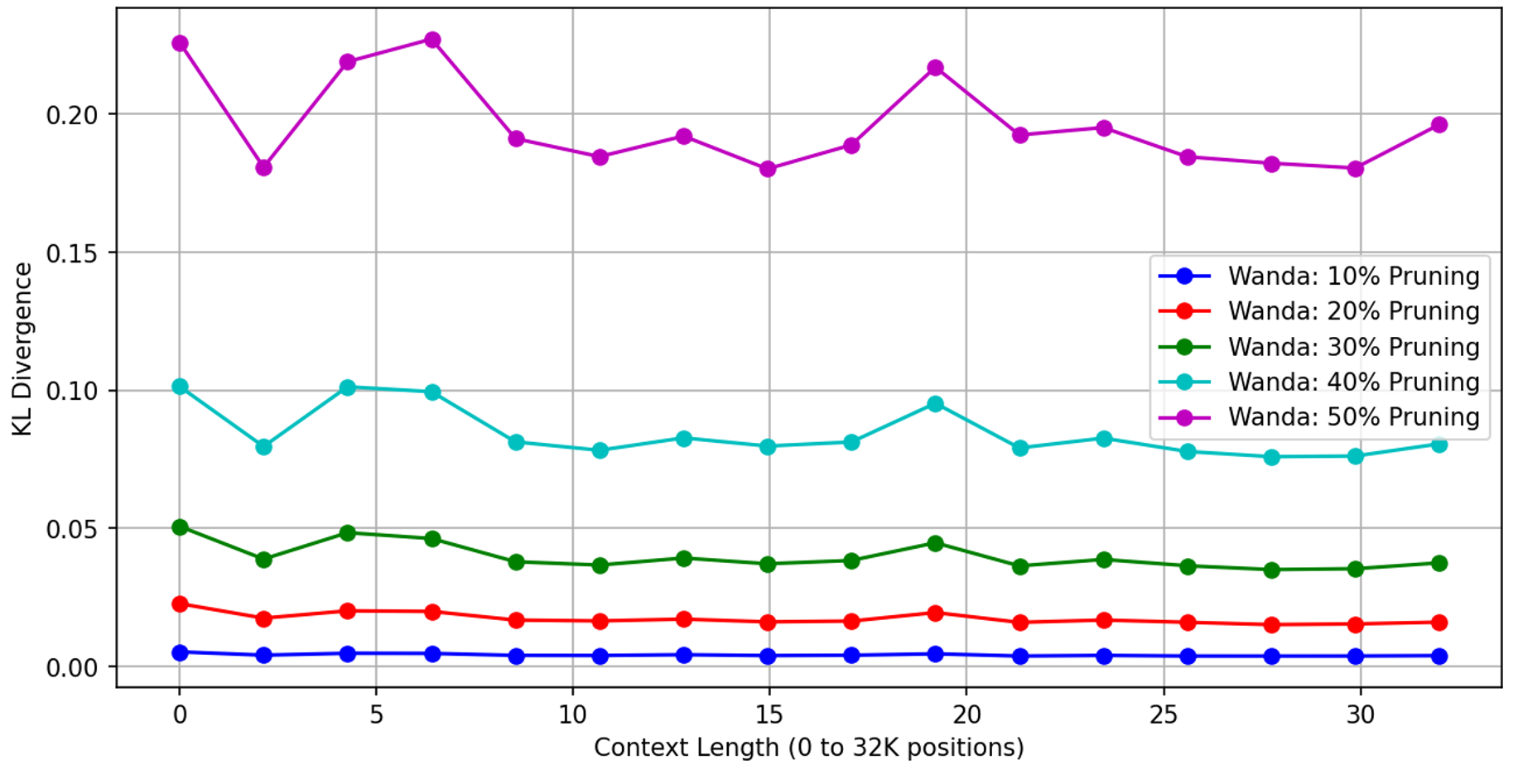}
     \end{subfigure}
    \caption{Pruning algorithms are robust to context lengths. The KL divergence of output logits between the uncompressed model and the pruned models does not change much with respect to different context lengths. The pruning ratio will only affect the variance of KL divergence values measured in different context lengths.}
    \label{results:prune}
\end{figure*}

This formula indicates that for compressed LLMs, the computation of the $t$-th hidden state incorporates not only the intended values but also cumulative noise from both key and value vectors. \textbf{The variance of the noise sum grows linearly with $t$, indicating increased computation error in longer sequences.}

However, due to the complexity of LLMs, there remains uncertainty regarding how this theoretical analysis will be reflected in the final output. In the following sub-section, we will discuss the empirical evaluation of the computation error of compressed LLMs under long-context. 

\subsection{Empirical Evaluation}
\label{sec:emprical}
To assess the computational error in the final output, we compute the Kullback-Leibler (KL) divergence between the outputs of the compressed and uncompressed versions of the same model, as indicated below:
\begin{align}
    D_{\text{KL}}(p \parallel q) &= \sum_{x} p(x) \log\left(\frac{p(x)}{q(x)}\right)
\end{align}
where $q$ represents the uncompressed model's output and $p$ represents the compressed model's output.

We have selected LLaMA-2-7B-32K as our model to evaluate four distinct LLM compression techniques across two categories\cite{LLaMA2-7B-32K}. For pruning, we implement magnitude pruning, which represents the simplest pruning technique, and Wanda pruning, which represents the state-of-the-art method. For quantization, we choose to evaluate weight-only and weight-activation quantization implemented by~\cite{li2024evaluating}. Due to its utilization of 'per-group' weight quantization and 'per-token' activation quantization, the method described in~\cite{li2024evaluating} represents a robust quantization technique. It achieves minimal computational error compared to other methods. We sampled texts from WikiText dataset~\cite{merity2016pointer}, and calculate the KL divergence between the models' output.

\begin{figure*}[h]
    \centering
    \begin{subfigure}[b]{0.45\textwidth}
         \centering
         \includegraphics[width=\textwidth]{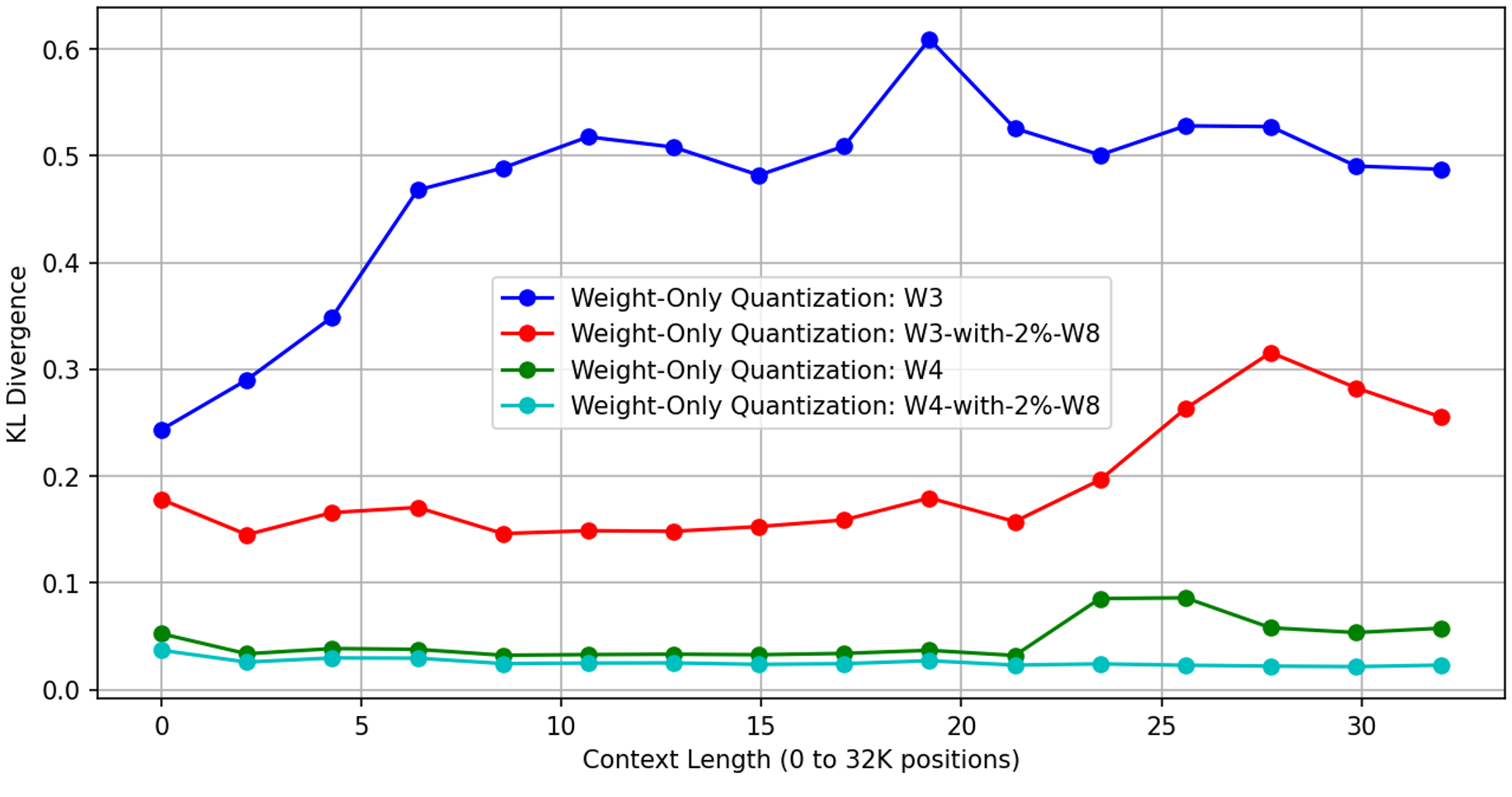}
    \end{subfigure}%
    ~ 
    \begin{subfigure}[t]{0.45\textwidth}
         \centering
         \includegraphics[width=\textwidth]{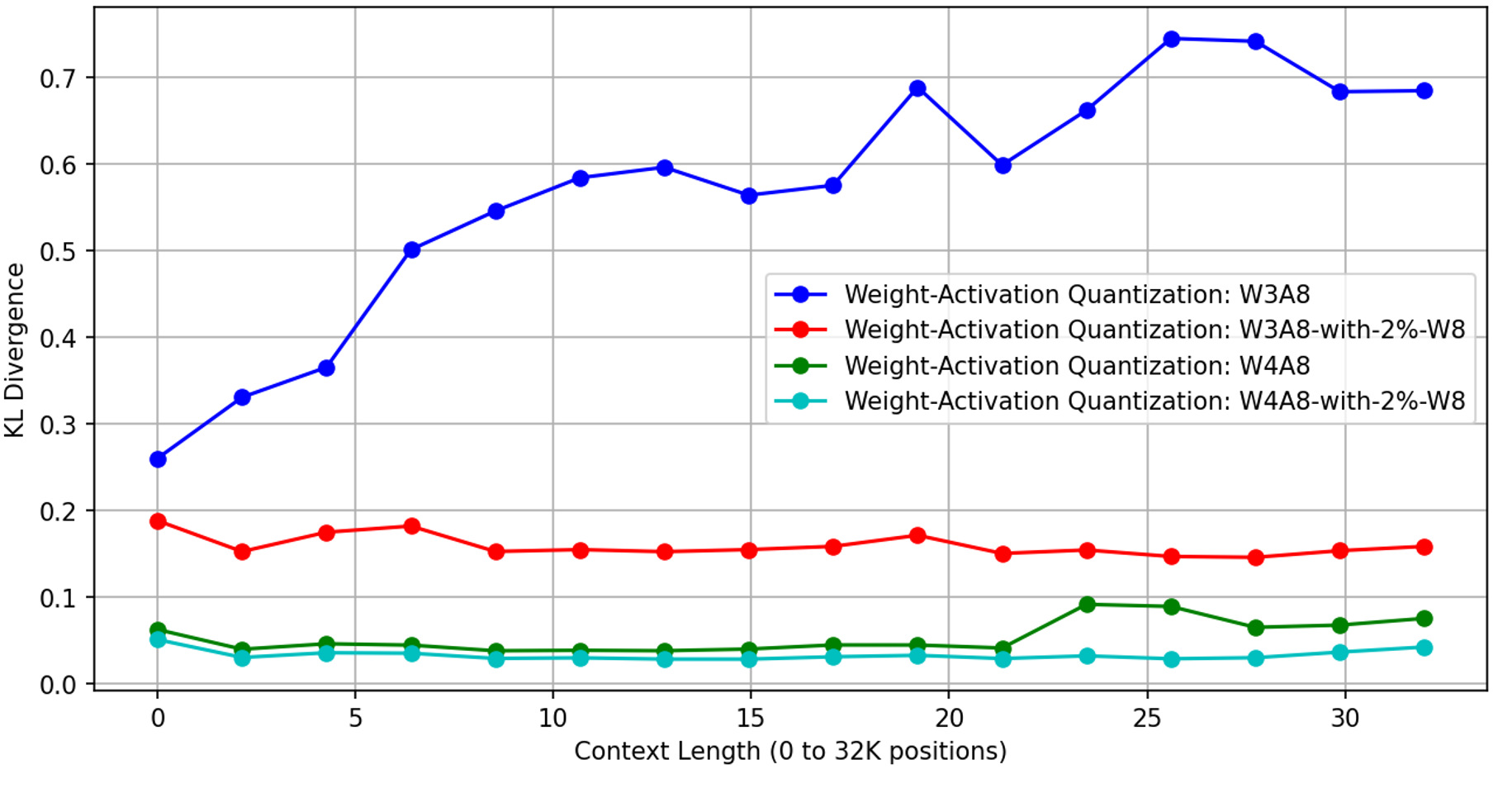}
     \end{subfigure}
    \caption{Only about 2\% weights are sensitive to low-bit quantizations. We can use 8-bit quantization instead of 3/4-bit quantization for these weights to make the compressed models less sensitive to context lengths.}
    \label{fig:2percent}
\end{figure*}

\begin{figure*}[t!]
    \centering
    \begin{subfigure}[b]{0.5\textwidth}
         \centering
         \includegraphics[width=\textwidth]{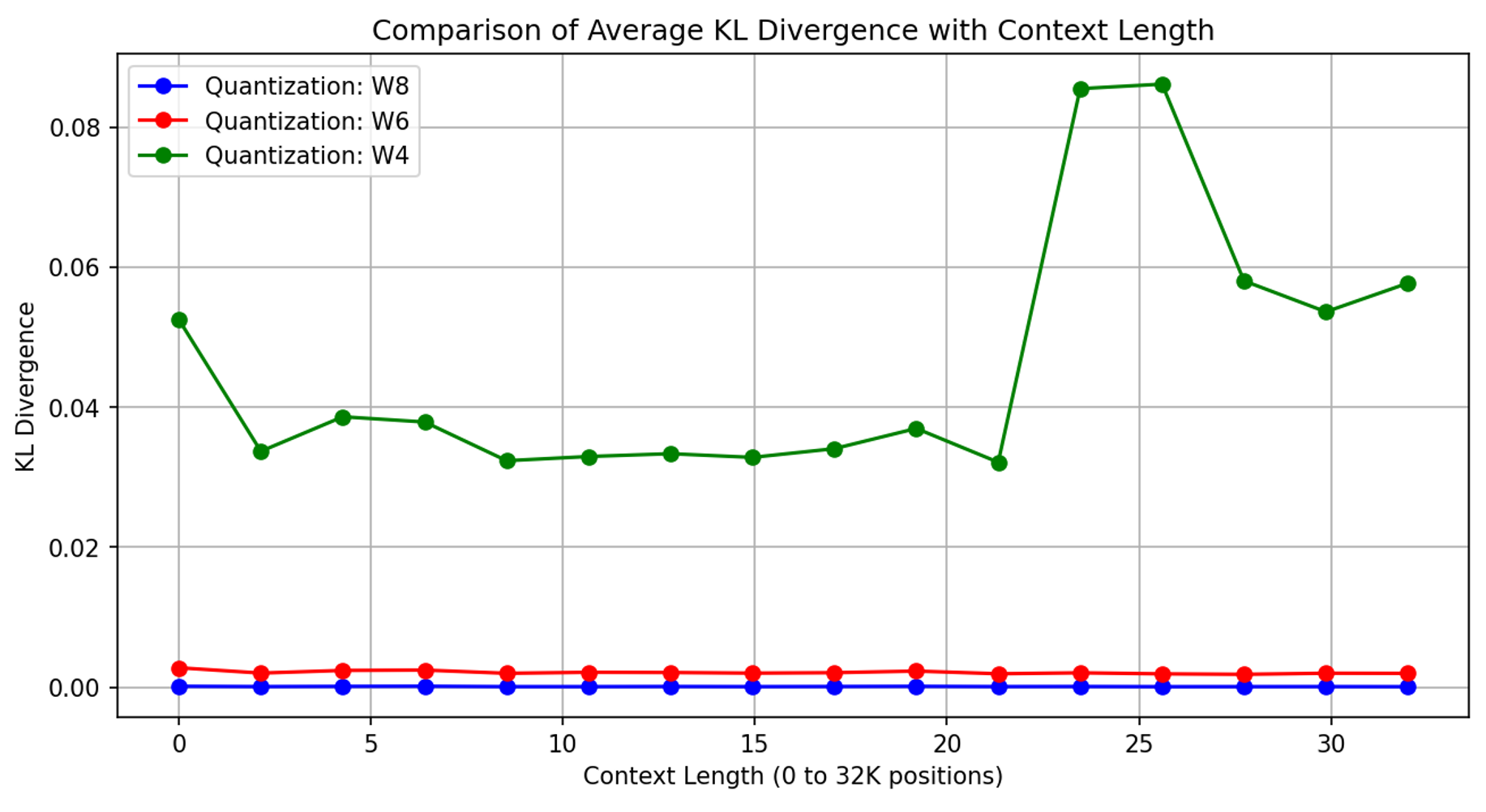}
    \end{subfigure}%
    ~ 
    \begin{subfigure}[t]{0.5\textwidth}
         \centering
         \includegraphics[width=\textwidth]{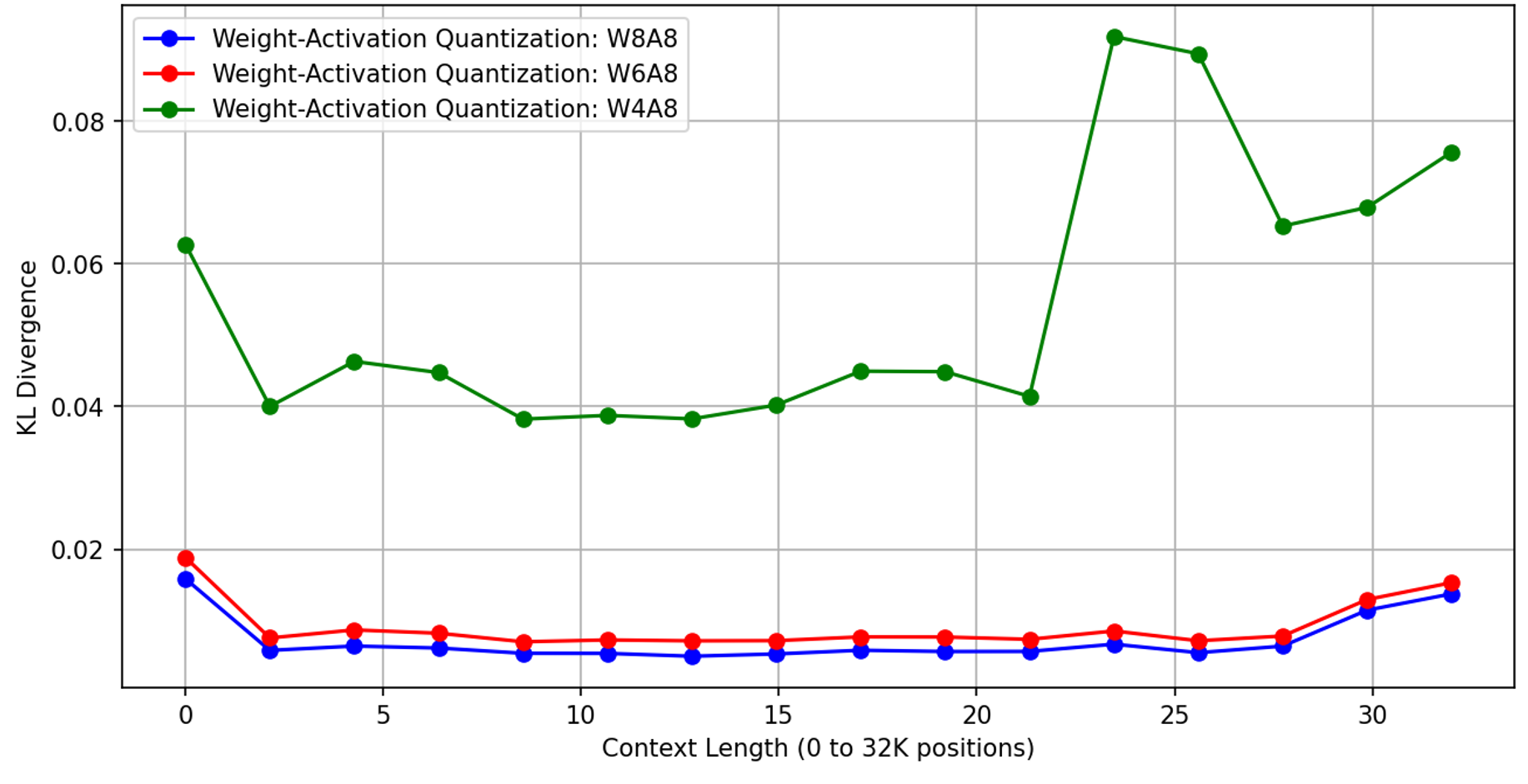}
     \end{subfigure}
    \caption{When we use low-bit ($\le 4$) weight quantization, the performance of compressed models becomes more sensitive to context lengths: the output of compressed models become more different from the of the uncompressed model when the context length increases.}
    \label{results:quant4}
\end{figure*}

\begin{figure}
    \centering
    \includegraphics[width=0.5\textwidth]{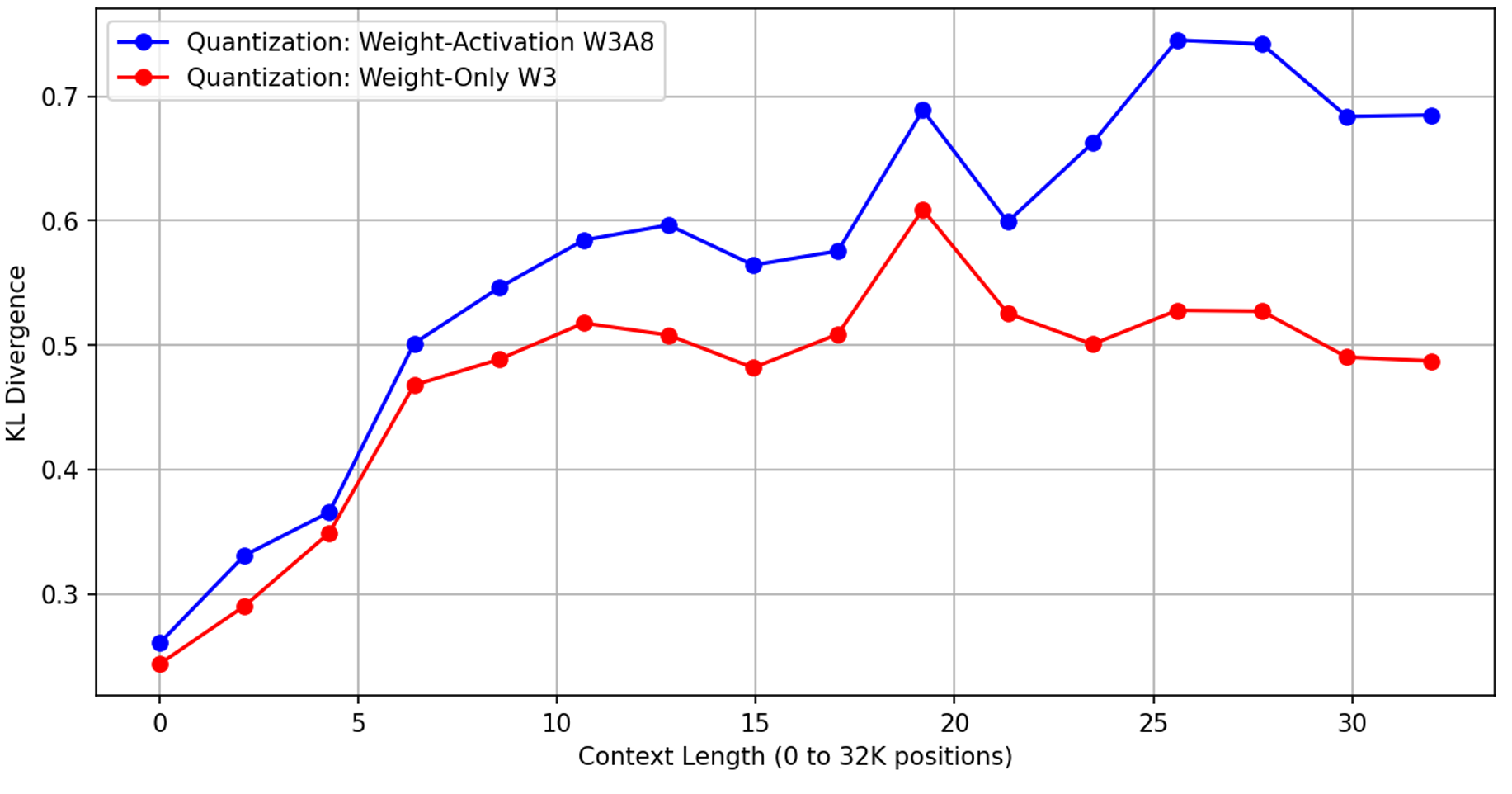}
    \caption{When choosing 3-bit quantization, it is very obvious that the output of compressed models become more different from the of the uncompressed model when the context length increases.}
    \label{results:quant3}
\end{figure}

We plot out experiments using pruning in Figure~\ref{results:prune} and experiments using quantization in Figure~\ref{results:quant4} and Figure~\ref{results:quant3}. \textbf{Counter-intuitively, the performance of pruning and quantization varies.} For pruning, the KL-divergence of output between the uncompressed model and the pruned models does not change much with respect to different context lengths. The pruning ratio will only affect the variance of KL-divergence values measured in different context lengths. For quantization, especially when we use low-bit ($\le 4$) weight quantization, the performance of compressed models becomes more sensitive to context lengths: the output of compressed models becomes more different from that of the uncompressed model as the context length increases.

\subsection{Hypothesis on the Varied Behaviors}

In this section, we attempt to discuss the reason why pruning remains robust as context length increases while quantization suffers from increasing computational error.

Our intuition is that for different pruning methods, the weights with larger magnitudes are always left untouched, while for quantization methods, they typically quantize all the weights regardless of their magnitude. As we observed in Section~\ref{sec:emprical}, pruning stays robust to context length. Thus, it is reasonable to hypothesize that \textbf{LLMs' long-range dependency only relies on a small part of weights, and for LLaMA-2-7B-32K, the weights with larger magnitude are more sensitive in longer contexts.}

Our experimental results support our intuition. As is shown in Figure~\ref{fig:2percent}, if we select about 2\% of weight groups with large magnitude, and quantize them to 8 bits, while quantizing other groups to 3 or 4 bits; we observe that for both weight-only quantization and weight-activation quantization, the increasing KL divergence under long-context disappears. However, the overall KL divergence level doesn't change much. In this way, we remedy the performance drop of low-bit quantized LLMs.

\begin{figure}
    \centering
    \includegraphics[width=0.5\textwidth]{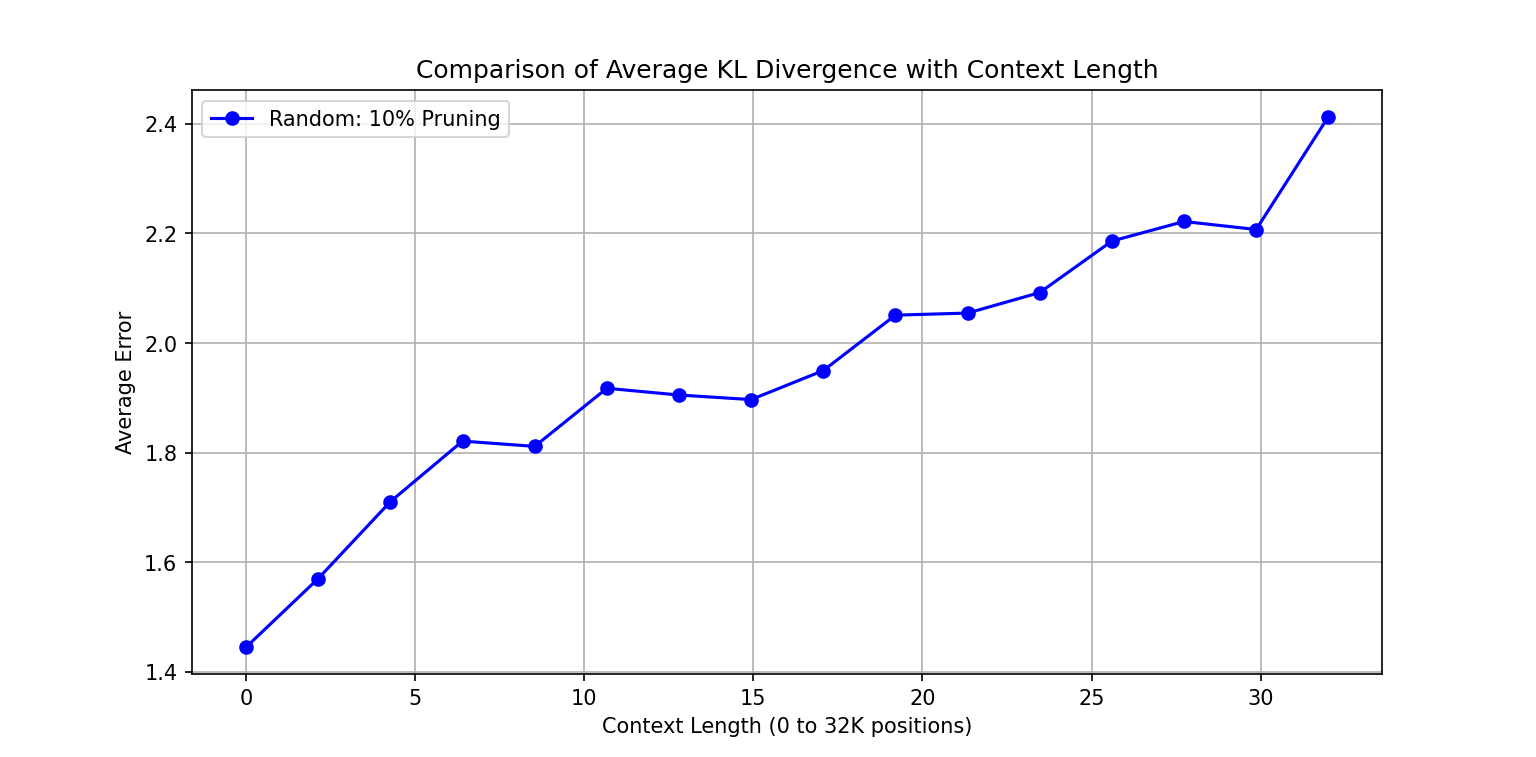}
    \caption{If we randomly prune out only 10\% weights, we can observe an almost-linearly increasing KL divergence as context length increases.}
    \label{results:random}
\end{figure}

Another experiment results to support our hypothesis is that for pruning, as is shown in Figure~\ref{results:random}, if we randomly prune out only 10\% of the weights, we can observe an almost linearly increasing KL divergence as context length increases.

\section{Conclusion}
\label{sec:conclusion}
This work explored the computation errors in zero-shot LLM compression within long-context settings, uncovering that computation errors escalate with increased context length in certain compression settings, while remaining stable in others. Our hypothesis suggests that specific weights are particularly sensitive to compression under long-context scenarios, highlighting a critical area for future research to optimize LLM performance effectively.

\section{Future Work}

In this project, we explore this question within a relatively simple setting. In the future, we plan to evaluate additional long-context LLMs, such as Llama 3 and Mistral \cite{zhang2024extending, jiang2023mistral}, and conduct experiments across a broader range of tasks. Additionally, we acknowledge that our hypothesis requires further validation. We aim to deepen our research to solidify these initial findings. We can also summarize our research as a new LLM compression methods if possible.

\section{Acknowledgements}

We extend our heartfelt gratitude to Prof. Kai Li and the teaching assistants for their invaluable guidance and support throughout this project. We also express our appreciation to Hao Kang, a PhD student at the Georgia Institute of Technology, for his insightful discussions that enriched our work. Chenyu is also grateful for Prof. Xuefei Ning and Shiyao Li from Tsinghua University for their work on LLM quantization evaluation. 

\newpage
{ \bibliographystyle{acm}
\bibliography{sample}}

\begin{thebibliography}{10}

\bibitem{longbench}
{\sc Bai, Y., Lv, X., Zhang, J., Lyu, H., Tang, J., Huang, Z., Du, Z., Liu, X., Zeng, A., Hou, L., et~al.}
\newblock Longbench: A bilingual, multitask benchmark for long context understanding.
\newblock {\em arXiv preprint arXiv:2308.14508\/} (2023).

\bibitem{black2022gpt}
{\sc Black, S., Biderman, S., Hallahan, E., Anthony, Q., Gao, L., Golding, L., He, H., Leahy, C., McDonell, K., Phang, J., et~al.}
\newblock Gpt-neox-20b: An open-source autoregressive language model.
\newblock {\em arXiv preprint arXiv:2204.06745\/} (2022).

\bibitem{bondarenko2021understanding}
{\sc Bondarenko, Y., Nagel, M., and Blankevoort, T.}
\newblock Understanding and overcoming the challenges of efficient transformer quantization.
\newblock In {\em Proceedings of the 2021 Conference on Empirical Methods in Natural Language Processing\/} (Online and Punta Cana, Dominican Republic, Nov. 2021), Association for Computational Linguistics, pp.~7947--7969.

\bibitem{gpt3}
{\sc Brown, T., Mann, B., Ryder, N., Subbiah, M., Kaplan, J.~D., Dhariwal, P., Neelakantan, A., Shyam, P., Sastry, G., Askell, A., Agarwal, S., Herbert-Voss, A., Krueger, G., Henighan, T., Child, R., Ramesh, A., Ziegler, D., Wu, J., Winter, C., Hesse, C., Chen, M., Sigler, E., Litwin, M., Gray, S., Chess, B., Clark, J., Berner, C., McCandlish, S., Radford, A., Sutskever, I., and Amodei, D.}
\newblock Language models are few-shot learners.
\newblock In {\em Advances in Neural Information Processing Systems\/} (2020), H.~Larochelle, M.~Ranzato, R.~Hadsell, M.~Balcan, and H.~Lin, Eds., vol.~33, Curran Associates, Inc., pp.~1877--1901.

\bibitem{cai2022deepguiser}
{\sc Cai, Y., Wang, C., Ning, X., Zhou, Z., Niu, D., Yang, H., and Wang, Y.}
\newblock Deepguiser: Learning to disguise neural architectures for impeding adversarial transfer attacks.

\bibitem{chen2023extending}
{\sc Chen, S., Wong, S., Chen, L., and Tian, Y.}
\newblock Extending context window of large language models via positional interpolation.
\newblock {\em arXiv preprint arXiv: 2306.15595\/} (2023).

\bibitem{chen2024longlora}
{\sc Chen, Y., Qian, S., Tang, H., Lai, X., Liu, Z., Han, S., and Jia, J.}
\newblock Longlo{RA}: Efficient fine-tuning of long-context large language models.
\newblock In {\em The Twelfth International Conference on Learning Representations\/} (2024).

\bibitem{LLaMA2-7B-32K}
{\sc Computer, T.}
\newblock {LLaMA-2-7B-32K}.
\newblock \url{https://huggingface.co/togethercomputer/LLaMA-2-7B-32K}, 2024.
\newblock Accessed: yyyy-mm-dd.

\bibitem{dettmers2022llmint8}
{\sc Dettmers, T., Lewis, M., Belkada, Y., and Zettlemoyer, L.}
\newblock Llm.int8(): 8-bit matrix multiplication for transformers at scale.
\newblock {\em arXiv preprint arXiv:2208.07339\/} (2022).

\bibitem{dettmers2022case}
{\sc Dettmers, T., and Zettlemoyer, L.}
\newblock The case for 4-bit precision: k-bit inference scaling laws.
\newblock {\em arXiv preprint arXiv:2212.09720\/} (2022).

\bibitem{frankle2018lottery}
{\sc Frankle, J., and Carbin, M.}
\newblock The lottery ticket hypothesis: Finding sparse, trainable neural networks.
\newblock {\em arXiv preprint arXiv:1803.03635\/} (2018).

\bibitem{frantar2022gptq}
{\sc Frantar, E., Ashkboos, S., Hoefler, T., and Alistarh, D.}
\newblock Gptq: Accurate post-training quantization for generative pre-trained transformers.
\newblock {\em arXiv preprint arXiv:2210.17323\/} (2022).

\bibitem{han2016deep}
{\sc Han, S., Mao, H., and Dally, W.~J.}
\newblock {Deep Compression: Compressing Deep Neural Networks with Pruning, Trained Quantization and Huffman Coding}.
\newblock In {\em ICLR\/} (2016).

\bibitem{han2015learning}
{\sc Han, S., Pool, J., Tran, J., and Dally, W.}
\newblock Learning both weights and connections for efficient neural network.
\newblock {\em Advances in neural information processing systems 28\/} (2015).

\bibitem{hoffmann2022training}
{\sc Hoffmann, J., Borgeaud, S., Mensch, A., Buchatskaya, E., Cai, T., Rutherford, E., Casas, D. d.~L., Hendricks, L.~A., Welbl, J., Clark, A., et~al.}
\newblock Training compute-optimal large language models.
\newblock {\em arXiv preprint arXiv:2203.15556\/} (2022).

\bibitem{jiang2023mistral}
{\sc Jiang, A.~Q., Sablayrolles, A., Mensch, A., Bamford, C., Chaplot, D.~S., Casas, D. d.~l., Bressand, F., Lengyel, G., Lample, G., Saulnier, L., et~al.}
\newblock Mistral 7b.
\newblock {\em arXiv preprint arXiv:2310.06825\/} (2023).

\bibitem{jiang_longllmlingua_2023}
{\sc Jiang, H., Wu, Q., Luo, X., Li, D., Lin, C.-Y., Yang, Y., and Qiu, L.}
\newblock {LongLLMLingua}: {Accelerating} and {Enhancing} {LLMs} in {Long} {Context} {Scenarios} via {Prompt} {Compression}, Oct. 2023.
\newblock arXiv:2310.06839 [cs].

\bibitem{kang2024gear}
{\sc Kang, H., Zhang, Q., Kundu, S., Jeong, G., Liu, Z., Krishna, T., and Zhao, T.}
\newblock Gear: An efficient kv cache compression recipefor near-lossless generative inference of llm.
\newblock {\em arXiv preprint arXiv:2403.05527\/} (2024).

\bibitem{kurtic2022optimal}
{\sc Kurtic, E., Campos, D., Nguyen, T., Frantar, E., Kurtz, M., Fineran, B., Goin, M., and Alistarh, D.}
\newblock The optimal bert surgeon: Scalable and accurate second-order pruning for large language models.
\newblock {\em arXiv preprint arXiv:2203.07259\/} (2022).

\bibitem{kwon2023efficient}
{\sc Kwon, W., Li, Z., Zhuang, S., Sheng, Y., Zheng, L., Yu, C.~H., Gonzalez, J., Zhang, H., and Stoica, I.}
\newblock Efficient memory management for large language model serving with pagedattention.
\newblock In {\em Proceedings of the 29th Symposium on Operating Systems Principles\/} (2023), pp.~611--626.

\bibitem{lecun1989optimal}
{\sc LeCun, Y., Denker, J., and Solla, S.}
\newblock Optimal brain damage.
\newblock {\em Advances in neural information processing systems 2\/} (1989).

\bibitem{li2024evaluating}
{\sc Li, S., Ning, X., Wang, L., Liu, T., Shi, X., Yan, S., Dai, G., Yang, H., and Wang, Y.}
\newblock Evaluating quantized large language models.
\newblock {\em arXiv preprint arXiv:2402.18158\/} (2024).

\bibitem{liu2019roberta}
{\sc Liu, Y., Ott, M., Goyal, N., Du, J., Joshi, M., Chen, D., Levy, O., Lewis, M., Zettlemoyer, L., and Stoyanov, V.}
\newblock Roberta: A robustly optimized bert pretraining approach.
\newblock {\em arXiv preprint arXiv:1907.11692\/} (2019).

\bibitem{lu2022learn}
{\sc Lu, P., Mishra, S., Xia, T., Qiu, L., Chang, K.-W., Zhu, S.-C., Tafjord, O., Clark, P., and Kalyan, A.}
\newblock Learn to explain: Multimodal reasoning via thought chains for science question answering.
\newblock {\em Advances in Neural Information Processing Systems 35\/} (2022), 2507--2521.

\bibitem{merity2016pointer}
{\sc Merity, S., Xiong, C., Bradbury, J., and Socher, R.}
\newblock Pointer sentinel mixture models, 2016.

\bibitem{openai2023gpt4}
{\sc OpenAI}.
\newblock Gpt-4 technical report, 2023.

\bibitem{park2022nuqmm}
{\sc Park, G., Park, B., Kwon, S.~J., Kim, B., Lee, Y., and Lee, D.}
\newblock nuqmm: Quantized matmul for efficient inference of large-scale generative language models.
\newblock {\em arXiv preprint arXiv:2206.09557\/} (2022).

\bibitem{paul2022unmasking}
{\sc Paul, M., Chen, F., Larsen, B.~W., Frankle, J., Ganguli, S., and Dziugaite, G.~K.}
\newblock Unmasking the lottery ticket hypothesis: What's encoded in a winning ticket's mask?
\newblock {\em arXiv preprint arXiv:2210.03044\/} (2022).

\bibitem{peng2024yarn}
{\sc Peng, B., Quesnelle, J., Fan, H., and Shippole, E.}
\newblock Ya{RN}: Efficient context window extension of large language models.
\newblock In {\em The Twelfth International Conference on Learning Representations\/} (2024).

\bibitem{sanh2021multitask}
{\sc Sanh, V., Webson, A., Raffel, C., Bach, S.~H., Sutawika, L., Alyafeai, Z., Chaffin, A., Stiegler, A., Scao, T.~L., Raja, A., et~al.}
\newblock Multitask prompted training enables zero-shot task generalization.
\newblock {\em arXiv preprint arXiv:2110.08207\/} (2021).

\bibitem{sheng_s-lora_2023}
{\sc Sheng, Y., Cao, S., Li, D., Hooper, C., Lee, N., Yang, S., Chou, C., Zhu, B., Zheng, L., Keutzer, K., Gonzalez, J.~E., and Stoica, I.}
\newblock S-{LoRA}: {Serving} {Thousands} of {Concurrent} {LoRA} {Adapters}, Nov. 2023.
\newblock arXiv:2311.03285 [cs].

\bibitem{sheng2023high}
{\sc Sheng, Y., Zheng, L., Yuan, B., Li, Z., Ryabinin, M., Fu, D.~Y., Xie, Z., Chen, B., Barrett, C., Gonzalez, J.~E., et~al.}
\newblock High-throughput generative inference of large language models with a single gpu.
\newblock {\em arXiv preprint arXiv:2303.06865\/} (2023).

\bibitem{su2021roformer}
{\sc Su, J., Lu, Y., Pan, S., Wen, B., and Liu, Y.}
\newblock Roformer: Enhanced transformer with rotary position embedding.
\newblock {\em NEUROCOMPUTING\/} (2021).

\bibitem{sun2022gibbon}
{\sc Sun, H., Wang, C., Zhu, Z., Ning, X., Dai, G., Yang, H., and Wang, Y.}
\newblock Gibbon: Efficient co-exploration of nn model and processing-in-memory architecture.
\newblock In {\em 2022 Design, Automation \& Test in Europe Conference \& Exhibition (DATE)\/} (2022), IEEE, pp.~867--872.

\bibitem{sun2023simple}
{\sc Sun, M., Liu, Z., Bair, A., and Kolter, J.~Z.}
\newblock A simple and effective pruning approach for large language models.
\newblock {\em arXiv preprint arXiv:2306.11695\/} (2023).

\bibitem{touvron2023llama2}
{\sc Touvron, H., Martin, L., Stone, K., Albert, P., Almahairi, A., Babaei, Y., Bashlykov, N., Batra, S., Bhargava, P., Bhosale, S., et~al.}
\newblock Llama 2: Open foundation and fine-tuned chat models.
\newblock {\em arXiv preprint arXiv:2307.09288\/} (2023).

\bibitem{wang2023epim}
{\sc Wang, C., Dong, Z., Zhou, D., Zhu, Z., Wang, Y., Feng, J., and Keutzer, K.}
\newblock Epim: Efficient processing-in-memory accelerators based on epitome.
\newblock {\em arXiv preprint arXiv:2311.07620\/} (2023).

\bibitem{wang2019eigendamage}
{\sc Wang, C., Grosse, R., Fidler, S., and Zhang, G.}
\newblock Eigendamage: Structured pruning in the kronecker-factored eigenbasis.
\newblock In {\em International conference on machine learning\/} (2019), PMLR, pp.~6566--6575.

\bibitem{wei2023outlier}
{\sc Wei, X., Zhang, Y., Li, Y., Zhang, X., Gong, R., Guo, J., and Liu, X.}
\newblock Outlier suppression+: Accurate quantization of large language models by equivalent and optimal shifting and scaling.
\newblock {\em arXiv preprint arXiv:2304.09145\/} (2023).

\bibitem{wei2022outlier}
{\sc Wei, X., Zhang, Y., Zhang, X., Gong, R., Zhang, S., Zhang, Q., Yu, F., and Liu, X.}
\newblock Outlier suppression: Pushing the limit of low-bit transformer language models.
\newblock {\em arXiv preprint arXiv:2209.13325\/} (2022).

\bibitem{xiao2022smoothquant}
{\sc Xiao, G., Lin, J., Seznec, M., Demouth, J., and Han, S.}
\newblock Smoothquant: Accurate and efficient post-training quantization for large language models.
\newblock {\em arXiv preprint arXiv:2211.10438\/} (2022).

\bibitem{zafrir2021prune}
{\sc Zafrir, O., Larey, A., Boudoukh, G., Shen, H., and Wasserblat, M.}
\newblock Prune once for all: Sparse pre-trained language models.
\newblock {\em arXiv preprint arXiv:2111.05754\/} (2021).

\bibitem{zhang2024extending}
{\sc Zhang, P., Shao, N., Liu, Z., Xiao, S., Qian, H., Ye, Q., and Dou, Z.}
\newblock Extending llama-3's context ten-fold overnight.
\newblock {\em arXiv preprint arXiv:2404.19553\/} (2024).

\bibitem{zhang2023llama}
{\sc Zhang, R., Han, J., Zhou, A., Hu, X., Yan, S., Lu, P., Li, H., Gao, P., and Qiao, Y.}
\newblock Llama-adapter: Efficient fine-tuning of language models with zero-init attention.
\newblock {\em arXiv preprint arXiv:2303.16199\/} (2023).

\bibitem{zhang2022opt}
{\sc Zhang, S., Roller, S., Goyal, N., Artetxe, M., Chen, M., Chen, S., Dewan, C., Diab, M., Li, X., Lin, X.~V., et~al.}
\newblock Opt: Open pre-trained transformer language models.
\newblock {\em arXiv preprint arXiv:2205.01068\/} (2022).

\bibitem{zoph2016neural}
{\sc Zoph, B., and Le, Q.~V.}
\newblock Neural architecture search with reinforcement learning.
\newblock {\em arXiv preprint arXiv:1611.01578\/} (2016).

\end{thebibliography}

\end{document}